\documentclass[a4paper, 10 pt, conference]{format/ieeeconf}   % Comment this line out if you need a4paper

\IEEEoverridecommandlockouts                              

\usepackage{graphicx}
\usepackage[table,dvipsnames]{xcolor} % colours & table support
\usepackage{amsmath,amssymb,amsfonts}
\usepackage{mathtools}                % extends amsmath
\usepackage{algorithm}
\usepackage[noend]{algpseudocode}     % modern algorithmicx interface
\usepackage{array,booktabs,multirow,eqparbox}
\usepackage[caption=false,font=footnotesize]{subfig}
\usepackage{tikz}
\usepackage{url,csquotes,pifont,tablefootnote,textcomp,scalerel}
\usepackage{siunitx}
\usepackage{balance}

\graphicspath{ {./figures/} }

\newcommand\copyrighttext{%
  \scriptsize\centering
  \textcopyright\ 2026 IEEE. Personal use of this material is permitted.
  Permission from IEEE must be obtained for all other uses, in any current or future
  media, including reprinting/republishing this material for advertising or promotional
  purposes, creating new collective works, for resale or redistribution to servers or
  lists, or reuse of any copyrighted component of this work in other works.\\
  Accepted manuscript. Published in the 2026 IEEE/SICE International Symposium on
  System Integration (SII), pp. 1259--1264.
  DOI: \url{https://doi.org/10.1109/SII64115.2026.11404550}.}
\newcommand\copyrightnotice{%
\begin{tikzpicture}[remember picture,overlay]
\node[anchor=south,yshift=10pt] at (current page.south) {\fbox{\parbox{\dimexpr\textwidth-\fboxsep-\fboxrule\relax}{\copyrighttext}}};
\end{tikzpicture}%
}

\title{\LARGE \bf
Towards Torque-Driven Reinforcement Learning for Quadruped Locomotion
}

\author{Jordan Dowdy$^{\text{~}1^*}$,~and Jean Chagas Vaz$^{\text{~}2}$
\thanks{$^{1}$J. Dowdy is a PhD student with the Department of Electrical and Computer Engineering at the University of Louisville.\break $^{*}$direct all correspondence to this author {\tt\small jordan.dowdy@louisville.edu}}
\thanks{$^{2}$Dr. Jean Chagas Vaz is with the Faculty of Electrical and Computer Engineering, University of Louisville, Louisville, KY, 40208, USA.\break  {\tt\small jean.chagasvaz@louisville.edu}}
}

\begin{document}

\maketitle
\thispagestyle{empty}
\pagestyle{empty}
\copyrightnotice

%%%%%%%%%%%%%%%%%%%%%%%%%%%%%%%%%%%%%%%%%%%%%%%%%%%%%%%%%%%%%%%%%%%%%%%%%%%%%%%%
\begin{abstract}

Reinforcement learning (RL) for legged robots is advancing locomotion, demonstrating its ability to adapt to new and challenging terrain. Traditionally, these RL locomotion frameworks are position-based, making the policy less adaptable to terrain types and requiring state estimation techniques in the observation space, i.e., linear velocity. Moreover, these RL frameworks often use small, lightweight quadrupeds that are limited in their viability for high-complexity tasks due to hardware constraints. This work explores an RL torque control framework for heavyweight high-torque quadrupeds. The RL framework in this paper can traverse rough terrain and effectively track a desired linear velocity without requiring knowledge of the agent's current velocity. Using Nvidia's Isaac Sim and Isaac Lab, simulation results of the RL torque control policy are shown on the Unitree B1 quadruped, achieving speeds of $3.5~m/s$ and $1.5~rad/s$. In addition, the quadruped can walk up and down stairs without the aid of an exteroceptive sensor.

\end{abstract}

%%%%%%%%%%%%%%%%%%%%%%%%%%%%%%%%%%%%%%%%%%%%%%%%%%%%%%%%%%%%%%%%%%%%%%%%%%%%%%%%
\section{INTRODUCTION}
Quadrupedal locomotion is a cornerstone of robotics research, alongside reinforcement learning, demonstrating its capability and robustness as a method for achieving efficient, adaptable legged locomotion. Quadrupeds have a distinct advantage over typical wheeled robots, as they can traverse challenging, rough terrain. This desirable capability presents an opportunity to apply quadrupeds in fields such as search and rescue, construction, and agriculture, where advanced locomotion is a baseline requirement for working in these industries. Larger quadrupeds, such as the Unitree B series and Boston Dynamics's Spot, show greater viability in these fields than their more agile counterparts, Unitree's Go lineup or MIT's mini-cheetah, as they demonstrate greater capability for traversing complex terrain while carrying payloads or equipment. These extra benefits increase the viability of these quadrupeds but introduce significant challenges, as the larger state space and highly nonlinear dynamics complicate the application of learning-based control frameworks.

\begin{figure}[t]
    \centering
    \setlength{\fboxsep}{0pt}%
    \setlength{\fboxrule}{1.5pt}%
    \fbox{\includegraphics[width=\dimexpr\linewidth-2\fboxrule\relax]{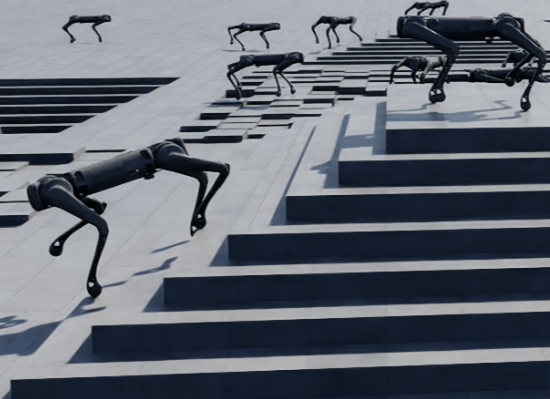}}
    \caption{\textbf{Unitree B1 walking down stairs in Isaac Sim} using a torque-driven RL locomotion controller.}
    \label{fig:B1inLab}
\end{figure}

Currently, legged locomotion is based on hybrid model control strategies to generate gaits and adapt to various terrains, utilizing algorithms such as Zero-Moment Point (ZMP) and model-predictive control (MPC)~\cite{10161260,8239554}. This approach uses a high-level controller to manage walking gaits and maintain stability, in conjunction with a low-level actuator controller. Although effective at achieving basic locomotion, this approach often fails to handle complex or challenging environments due to high-dimensional, nonlinear dynamics, especially in rugged terrain. In recent years, legged locomotion has seen increased interest in RL-based locomotion controllers. Reinforcement learning and simulation tools have demonstrated their ability to produce robust, adaptable learning-based control strategies~\cite{2109.11978} that often outperform MPC for high-level tasks~\cite{2501.16590}. Recent research highlights the superiority of torque-controlled RL frameworks, which allow direct actuator interaction to model actuator dynamics during training. These RL torque control frameworks have demonstrated enhanced dynamic stability and adaptability compared to position-based counterparts for quadrupeds in adverse terrain environments. The majority of quadruped locomotion research has focused on smaller quadrupeds, such as Unitree's GO1 and A1, which benefit from their compact size, low weight, and reduced state-space complexity, but fail to capture the challenges posed by larger quadrupeds like Unitree's B1 or Boston Dynamics' Spot. Larger quadrupeds face significantly increased system complexities, including higher inertia, greater contact forces, and the need to account for surface compliance, all of which increase the difficulty of both model training and policy design. This work aims to address these challenges by developing an RL-based locomotion framework tailored for larger quadrupeds while minimizing the complexity of the network architecture. Using torque-based control policies, we enable robust locomotion in adverse terrain environments, addressing the unique dynamics and control requirements of larger quadrupeds. 

\subsection*{Paper Contributions}

The locomotion controller's primary features, along with this paper's main contributions, are:
\begin{enumerate}
    \item A torque-driven whole-body reinforcement learning control policy.  
    \item Integration of the Unitree B1 into Nvidia's Isaac Sim.
    \item A discussion on the RL policy's architecture. 
    \item Preliminary policy results and experimentation in simulation.
\end{enumerate}
A discussion on our RL policy features, advantages, disadvantages, and implementation 
\subsection*{Article Structure}
This paper contains six sections: Section II reviews the literature on classical locomotion in quadrupeds and learning-based locomotion. Section III details our methodology and contains RL theory, policy architecture, and reward design choices. Section IV explains the experiment design and structure, while Section V discusses the experimental results. Lastly, Section VI concludes the paper with a reflection on the work and future improvements.

\section{Related Work}

\begin{figure*}[h!]
    \centering
    \hspace{1mm}
    \includegraphics[width=0.98\linewidth]{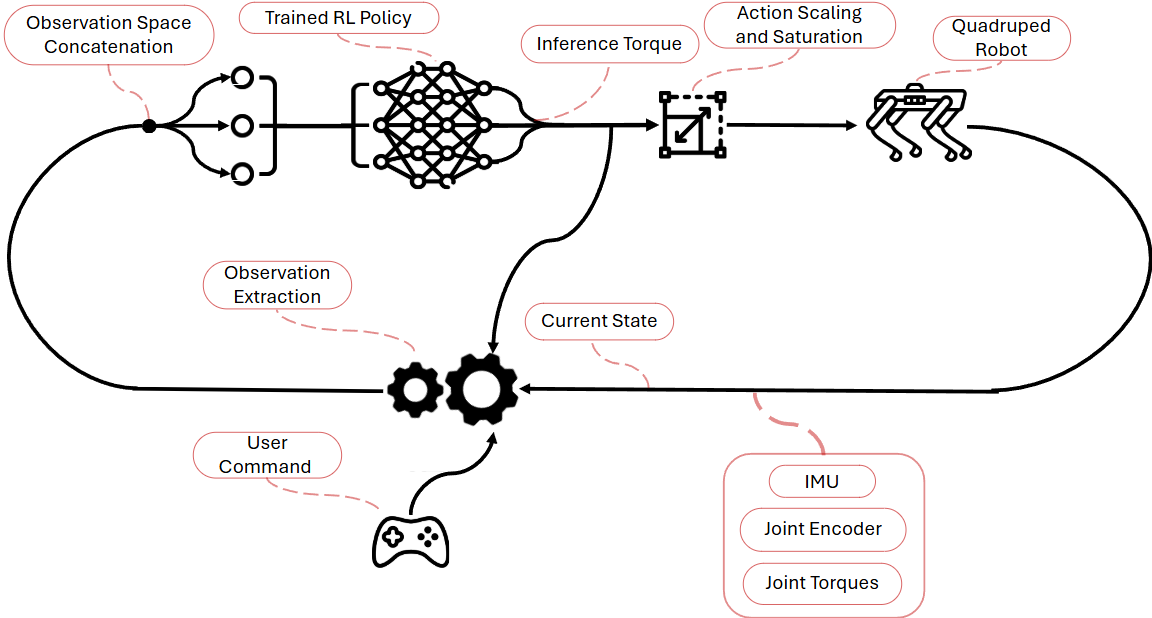}
    \caption{\textbf{Reinforcement learning policy control loop} after training. Current states are measured using their respective sensors: the Inertial Measurement Unit (IMU), joint encoders, and estimated joint torques. These states, along with the previous policy output and user commands, are combined to form a desired linear and angular body velocity and used during policy inference. The inference torques are scaled and clipped to within the actuator's capabilities and then provided to the quadruped robot.}
    \label{fig:RLCL}
\end{figure*}

The following subsections present relevant work by researchers, focusing on quadruped locomotion using classical and reinforcement learning approaches.

\subsubsection{Classical Quadrupedal Locomotion}
Legged locomotion continues to be widely studied and remains challenging to achieve, especially in non-ideal terrain. Most currently applied approaches employ a multi-modal or hierarchical control scheme for footstep planning, trajectory optimization, and Model Predictive Control (MPC)~\cite{10450998,10773113}. Recent advances have shown that learning-based techniques can achieve superior stability and complete more challenging tasks~\cite{2203.05194}. These RL frameworks are trained in simulation and then transferred to the physical robot. One of the significant issues in simulated environment training is the so-called sim-to-real gap~\cite{9606868}. The sim-to-real gap refers to a policy being robust enough to be applied in a real-world setting where some unmolded or unknown environmental factors are present, but absent in simulation training. The sim-to-real gap is often overcome by domain randomization, noise injection, and system disturbances~\cite{10343164}. Simulation training offers significant benefits in terms of training time and robustness, while also alleviating the need for physical hardware. While training can be conducted in simulation, several works have demonstrated the effectiveness of real-world training for locomotion policies~\cite {9812166,IsaacLab}. Often, these RL policies utilize an action space consisting of joint positions~\cite{9982132,2212.03238}. Recent success has been achieved using an action space of joint torques~\cite{2203.05194}, leading to improved locomotion quality. 

\subsubsection{Reinforcement Learning Quadrupedal Locomotion}
Reinforcement learning in quadruped research typically uses smaller, more cost-effective quadrupeds, such as Unitree's Go1 or AlienGo. While these small quadrupeds offer benefits due to their low barrier to entry and price tag, they are less suitable for use in challenging terrain and are less applicable to search-and-rescue or heavy payload-oriented tasks. Other quadrupeds, such as the Unitree B1 or ANYmal series, are less cost-effective. However, they accomplish tasks in these extremely adverse terrain environments by equipping them with high-torque actuators and a large frame. These quadrupeds' increased size and weight can make learning-based locomotion more challenging, as their system dynamics are more complex than those of cost-effective quadrupeds. Work on RL policies for these larger quadrupeds typically focuses on network architecture, using more complex architectures or policy optimization~\cite{9028178}. In contrast, this work focuses on reward and observation shaping to achieve locomotion, while maintaining a familiar network architecture often seen in RL locomotion research. As most RL policies in robotics train in simulation using Isaac Gym, Isaac Lab, or Mujoco, keeping the network architecture in line with what is pre-existing in these simulators, and with what is often used in related research, allows for this policy to be easily integrated and adapted for further development and completion of complex quadrupedal tasks.

\section{Reinforcement Learning Framework}
This section presents the methodology for training the Unitree B1 quadruped. It begins with a general outline of our simulation environment, followed by a formal problem definition. Then, details of the actions, observations, and rewards for the policy are discussed and compared with those of standard policies for smaller quadrupeds.

\subsection{Simulation Enviroment}
To simulate the Unitree B1 quadruped, we chose NVIDIA's Isaac Sim for environment and physics simulation, which uses PhysX, while using Isaac Lab~\cite{2109.11978} for our reinforcement learning workflow. An NVIDIA RTX 3090 GPU was used to run 4096 environments simultaneously for policy training. Domain randomization during training was used to ensure a robust policy that overcomes the sim-to-real gap, with foot friction, mass, initial joint positions, and initial velocities randomized by $\pm5$ to $20\%$ per term. Table~\ref{tab:DomRand} describes in more detail the terms and respective quantities used. Terrain difficulty gradually increased during training, accompanied by random pushing events that served as disturbances to the system. These random pushing events would occur every 10-15 seconds by setting the linear velocity to a value in the range~$(\text{-}0.5, 0.5)~m/s$. Command velocities are sampled from the ranges $(\text{-}2.0, 3.0)~m/s$ in X, $(\text{-}1.5,1.5)~m/s$ in Y, and $(\text{-}2.0, 2.0)~rad/s$ for linear and angular velocities, respectively. Fig.~\ref{fig:RLCL} shows and represents the inference loop after training. This inference loop remains the same for both simulation and the real world after training is complete, allowing the policy to be thoroughly tested in simulation before the sim-to-real transfer.  

\begin{table}\centering
\caption{Domain Randomization}\label{tab:DomRand}
\setlength{\tabcolsep}{4pt}
\renewcommand{\arraystretch}{1.5}
\begin{tabular}{cccl}
\toprule
\textbf{Term}              & \textbf{Operation} & \textbf{Range}\\ 
\midrule
Foot Friction  &  New    & $(0.4, 1.1)$
\\
Center of Mass    & Add   & $(\text{-}0.025,0.025)$                    \\
Link Mass  & Scale        & $(0.8,1.3)$         \\
Body Velocity & Add & $(\text{-}\hspace{0.1em}0.25, 0.25)$ \\
Body Position & Add & $(\text{-}\hspace{0.1em}0.5,0.5)$ \\
Body Orientation & Add & $(\text{-}\hspace{0.1em}0.02,0.02)$ \\
Joint Positions  & Scale  & $(\text{-}\hspace{0.1em}0.3, 0.3)$ \\
Joint Velocities  & Scale  & $(\text{-}\hspace{0.1em}2.5, 2.5)$ \\
\bottomrule
\end{tabular}    
\end{table}

\subsection{Problem Definition}
In reinforcement learning for the locomotion of a quadruped or agent, a Markov Decision Process (MDP) is used to model the robot's interaction with its environment. MDP consists of ($\mathcal{S}, \mathcal{A}, \mathcal{P}, \mathcal{R}, \gamma$), where $\mathcal{S}$ and $\mathcal{A}$ are the current state and possible future state representing the agent's state space. $\mathcal{P}$ is the transition probability which shows the relationship between actions $\mathcal{A}_t$ and the change in states at $\mathcal{S}_t$ to the next time step $\mathcal{S}_{t\text{+}1}$, $\mathcal{R}$ is the reward function representing the reward received after taking action $\mathcal{A}_t$ in state $\mathcal{S}_t$, finally, $\gamma$ represents the discount factor and determines the importance of future rewards. The agent's goal is to maximize the reward given by: 
\begin{equation}
    G_t = \sum^{\infty}_{k=0} \gamma^k \mathcal{R}(\mathcal{S}_{t\text{+}k},\mathcal{A}_{t\text{+}k}),\label{eq:CDR}
\end{equation}
Where a policy $\pi(\mathcal{A}|\mathcal{S}$ defines the probability of taking an action in a state. The expected return starting from a state $\mathcal{S}$ under policy $\pi$ is formulated as:
\begin{equation}
    V_{\pi} = \mathbb{E}_{\pi} \sum^{\infty}_{k=0} \gamma^k \mathcal{R}(\mathcal{S}_{t},\mathcal{A}_{t}|\mathcal{S}_0=\mathcal{S}),\label{eq:SVF}
\end{equation}
Similarly, the expected return starting from a state $\mathcal{S}$ taking action $\mathcal{A}$ under policy $\pi$ is formulated as: 

\begin{equation}
    Q_{\pi}(\mathcal{S},\mathcal{A}) = \mathbb{E}_{\pi} \sum^{\infty}_{k=0} \gamma^k \mathcal{R}(\mathcal{S}_{t},\mathcal{A}_{t}|\mathcal{S}_0=\mathcal{S},\mathcal{A}_0=\mathcal{A})\label{eq:AVF},
\end{equation}
Where the optimal policy $\pi^*$ maximizes the following value functions,
\begin{equation}
    \pi^* = \arg \max_{\pi}V^{\pi}(\mathcal{S}),
\end{equation}
\begin{equation}
    \pi^* = \arg \max_{\pi}Q^{\pi}(\mathcal{S},\mathcal{A}).
\end{equation}

\subsection{Action Space}
The action space for the policy is represented by a 12$\times$1 vector, corresponding to the joint torque of each actuator for the Unitree B1 quadruped. With a larger action space than one made of joint positions, policy actions are multiplied by a constant scaler ($30.0$). The purpose of this scaling is to simplify exploration within the quadruped's larger action space, as B1's actuator torque range is approximately $\pm140~Nm$, compared to traditional position-control RL frameworks with an action space of $\pm3.14~rad$. Given the policy's focus on locomotion, the need for the agent to explore higher torque values is less compelling; by setting the scale factor to a lower value, the agent is guided to explore smaller, stable, and task-relevant torque values early in training. This helps ensure initial stability by guiding the policy toward stable actions, thereby enabling more efficient exploration.

\subsection{Observations}
The observation space for our policy builds on traditional RL frameworks for torque and position control by introducing state information that is often omitted from the observation states. These added states aim to help the policy gain a deeper understanding of its system dynamics. The observation space comprises nine key terms: linear acceleration, angular velocity, Orientation, user command, joint positions, joint velocities, previous actuator output torque, previous policy action, and feet contact. Table~\ref{tab:obsTerms} gives a detailed look into these observations. Notably, linear velocity is not part of this policy's observation space, as a goal was to capture state relationships and eliminate the need for standard state estimation algorithms. Adding linear acceleration and the previous actuator output torque as states helps capture a few desired properties of RL policies, with the intuition that actuator dynamics can be learned by the policy when combined with domain randomization, and that it can more accurately capture tipping or collision events. The agent's orientation in the world frame is represented as a unit quaternion, replacing the projected gravity term often seen in other RL frameworks. User command represents the desired linear (X, Y) and angular (Z) velocities of the robot's base. The joint position observation uses a relative position, where $\theta_i$ denotes the desired stance the policy should maintain during training. The previous action term, provided in its unscaled form, is used to ensure the policy learns how its actions change its subsequent states during locomotion. At this moment in time, Isaac Lab does not allow access to applied torques on the actuator from the environment, such as the ground reaction force. The feet-contact observation was added to circumvent this and returns a binary signal when the quadruped's feet have collided.

\begin{table}\centering
\caption{Observation Space Terms}\label{tab:obsTerms}
\setlength{\tabcolsep}{4pt}
\renewcommand{\arraystretch}{1.3}
\begin{tabular}{ccccl}
\toprule
\textbf{Observation Term}               & \textbf{State Size}                                              & \textbf{Scaler}                  & \textbf{Injected Noise}    \\
\midrule
Linear Acceleration            &$ \textbf{a}_{base} \in \mathbb{R}^3$                    &0.7                     &0.02                      \\  
Angular Velocity                & $\boldsymbol{\omega}_{yaw}$ $\in \mathbb{R}^3$                    &0.7                     &0.01                       \\
Orientation               & $\textbf{O}_{base} \in \mathbb{R}^4$                    &0.7                    &0.05                      \\
User Command                   &$ \textbf{u}^*_{base}\in\mathbb{R}^3$                    &1            &0.0                       \\
Joint Positions                & \textbf{$\left|\left|\theta_i-\theta\right|\right|$} $\in \mathbb{R}^{12}$                   &1.0                     &0.05                      \\
Joint Velocity                 & \textbf{$\Dot{\theta}$} $\in \mathbb{R}^{12}$             &1                    &0.1                       \\
Previous Applied Torque      & \textbf{$\tau_{t-1}$} $\in \mathbb{R}^{12}$             &0.5                     &0.05                      \\
Previous Action                & \textbf{$\mathcal{A}_{t-1.}$} $\in \mathbb{R}^{12}$      &1.0                     &0.0                       \\
Feet Contact                   & \textbf{$C_{foot}$} $\in \mathbb{R}^4$                           &2.0                     &0.0                       \\
\bottomrule
\end{tabular}    
\end{table}

\subsection{Rewards}
We design the reward functions to achieve a stable periodic gate, maintain desired command velocities, and remain alive, defined as avoiding contact with the base, hip, and thighs.     

\begin{table}\centering
\caption{Reward and Penalty Terms}\label{tab:rwdTerms}
\setlength{\tabcolsep}{4pt}
\renewcommand{\arraystretch}{1.5}
\begin{tabular}{cccl}
\toprule
\textbf{Reward Term}                    & \textbf{General Expression}                                      & \textbf{Weight}       \\
\midrule
Alive                        & $\mathit{sign(alive)}$                                                       &0.75           \\  
Feet Air Time                & $\sum^4_{i=1}(t_{\mathit{air},i} - 0.3)$                              &5.0           \\
Linear Velocity Error        & $\mathit{exp}(-(||\upsilon_{x,y}^{\mathit{des}} - \upsilon_{x,y}^{\mathit{cur}}||/0.25) )$                    &7.0          \\
Angular Velocity Error       &$\mathit{exp}(-(||\omega_{\mathit{z}}^{\mathit{des}} - \omega_{\mathit{z}}^{\mathit{cur}}||/0.25) )$                    &5.0           \\
Gait                         & $\prod^2_{i=1}\ell_{\mathit{sync},i}\cdot\prod^4_{i=1}\ell_{\mathit{async},i}$                     &10.0  \\
\midrule
\textbf{Penalty Term}                   & \textbf{General Expression}                                           & \textbf{Weight}       \\
\midrule
Action Smoothness                & $||\mathcal{A}_t - \mathcal{A}_{t\text{-}\scaleto{1\mathstrut}{5.5pt}}||$                  &-0.6         \\
Air Time Variance                & $\sum^{4,4}_{i=1,j=1}(t_{\mathit{air},i} - t_{\mathit{air},j})$             &-1.0           \\
Base Motion                      & $| \omega_x \text{+} \omega_y |$      &-2.0         \\
Base Orientation                 & $||\textbf{G}_{x,y}||$                           &-3.0           \\
Foot Slippage                    & $||\textbf{C} \cdot \upsilon_{\mathit{foot}}||$                         &-0.5           \\
Joint Position Deviation              & $\sum^{12}_{h=1}(\theta_{\mathit{i}}-\theta)$                           &-0.7           \\

Joint Torque                     & $||\tau||$                          &-0.0005           \\
Joint Acceleration               & $||{\Ddot{\theta}}||$                           &-0.000025           \\
Joint Velocity                   & $||\dot{\theta}||$                           &-0.01           \\
Thigh Contact                          & \textbf{$C_{thigh}$} $\in \mathbb{R}^4$  &-1.0 \\
Calf Contact                          & \textbf{$C_{Calf}$} $\in \mathbb{R}^4$  &-0.2 \\
\bottomrule
\end{tabular}

\end{table}

\section{Network Architecture and Simulation Results}
With our torque control framework now detailed, the network architecture used to deploy this policy will be briefly summarized, and training results will be presented and discussed.

\begin{figure}[b!]
    \centering
    \vspace{-2mm}
    \includegraphics[width=0.99\linewidth]{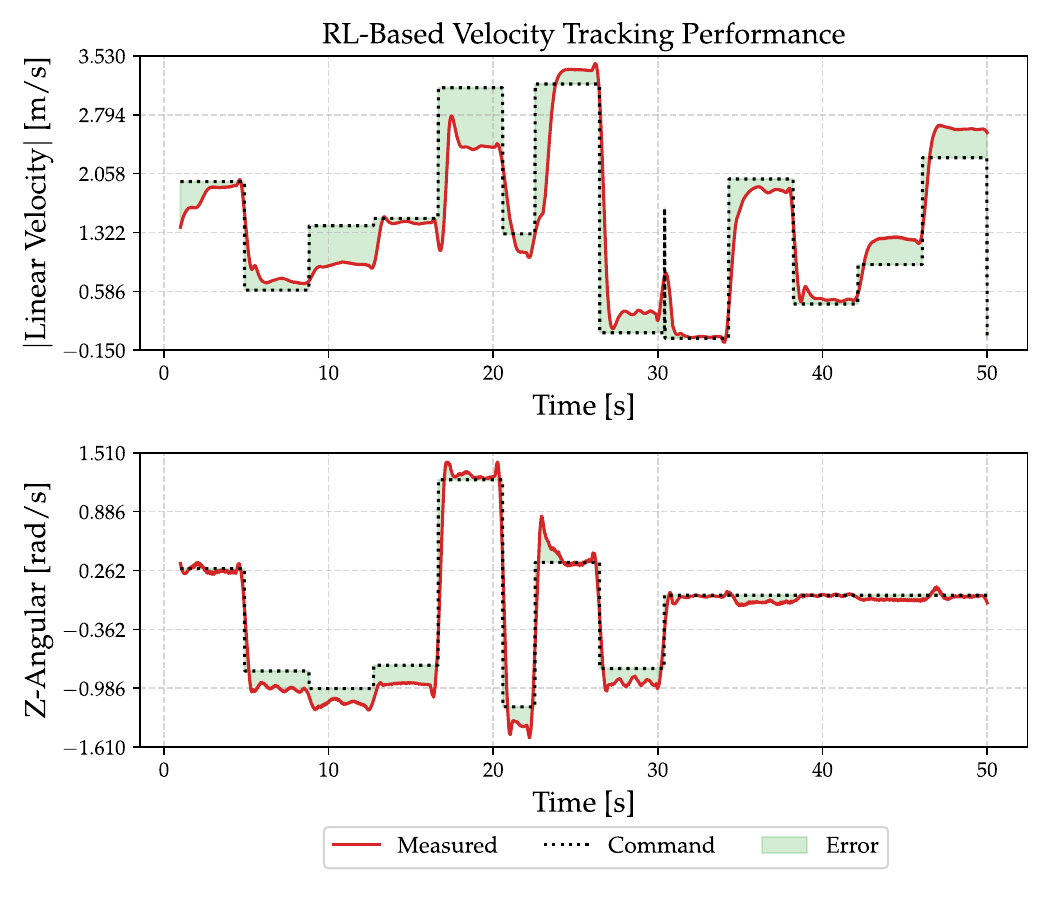}
    \caption{\textbf{Policy Velocity Tracking Performance} in Isaac Sim. Linear and angular velocity commands are randomly sampled for 30 seconds; afterward, pure linear velocity tracking performance is observed. The red curve shows the measured velocity, while the dashed curve represents the desired velocity. The green shading represents the error between the desired and measured velocity for the quadruped.}
    \vspace{1mm}
    \label{fig:simResults}
\end{figure}

\subsection{Network Architecture and Training}
The reinforcement learning policy in this work is modeled using a straightforward multilayer perceptron with three layers of 512, 256, and 128 neurons, employing exponential linear units (ELUs) as activation functions. Proximal policy optimization (PPO) is used to train the policy with tuned hyperparameters, including an adaptive learning rate, an entropy coefficient of 0.0025, a target KL-Divergence at 0.01, a discount factor ($\gamma$) and advantage estimator discount factor ($\lambda$) of $0.99$ and $0.95$, respectively, and a learning rate of 0.001. The policy training is conducted over 20,000 iterations, taking approximately 16 hours to complete. Fig.~\ref{fig:B1inLab} shows a simulation of the trained policy in Isaac Lab using the rough terrain environment.

The network architecture is intentionally kept simple in comparison to more complex actor-critic models or advanced optimization frameworks \cite{9028178}. This network architecture acquires a robust locomotion policy by focusing on reward and observation shaping rather than over-complicating the network architecture. Additionally, by leveraging Isaac Lab's out-of-the-box capabilities, we ensure the policy is modular and adaptable, allowing users to modify policy states for various quadruped locomotion tasks. This "plug and play" workflow enables users to experiment with diverse tasks and environments without overhauling the underlying RL framework, supporting the broader goal of a generalizable solution for large quadruped locomotion.

\subsection{Simulation Results}
\begin{figure}
    \centering
    \vspace{3mm}
    \includegraphics[width=3.325in,]{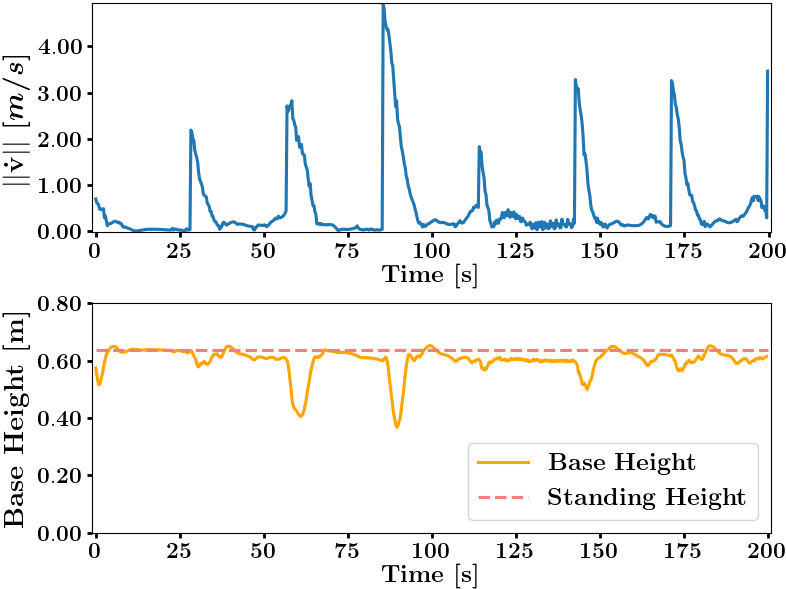}
    \caption{\textbf{Policy stability during pushing events.} The base of the robot was set to a random velocity in the range~$\pm5~m/s$ for both $x$ and $y$. The robot's base height was used as a benchmark to determine whether it could recover after a pushing event. The dashed red line represents the robot's height in its nominal configuration.}
    \label{fig:pushEvent}
\end{figure}

After policy training, velocity tracking performance was recorded for a single agent in Isaac Sim, with randomly sampled desired linear and angular velocities shown in Fig.~\ref{fig:simResults}. The red, dashed lines show the agent's actual and desired velocities, with green indicating the error. After 30 seconds, the angular velocity command was set to $0.0~rad/s$ to demonstrate the policy's performance when tracking only linear velocity. From Fig.~\ref{fig:simResults} (Right), the policy closely tracks the desired angular velocity. When large jumps in desired angular velocity occur, the policy takes approximately 0.5 seconds to come within $\pm0.2~rad/s$ of the newly sampled command. The linear velocity tracking performance of the policy shown in Fig.~\ref{fig:simResults} demonstrates that the model can capture state relationships to estimate its linear velocity and achieve velocities close to the desired ones. The robustness of the policy can be seen in Fig.~\ref{fig:pushEvent}, where the robot's base is set to a random linear velocity of $\pm5~m/s$ in both the $x$ and $y$ directions. Setting the instantaneous velocity of the base can represent a push being applied to the robot. The base height was used to determine if 1) the robot fell to the ground, and 2) if the policy was able to recover afterward. From Fig.~\ref{fig:pushEvent}, the policy can both resist and recover from these pushing events effectively. 

Visual inspection of the policy during simulation revealed artifacts due to a long iteration time. This overfitting manifested as a longer swing of each leg at low angular and linear velocities, likely because the policy attempts to maximize the feet's airtime reward while minimizing the airtime variance penalty at lower velocities. 

\section{Conclusion and Future Work}
In this paper, a reinforcement learning torque control framework was presented, aiming to improve the performance of high-inertia quadrupeds that have shown limited success in learning-based locomotion. This RL framework estimates the necessary joint torques to achieve stable locomotion while following a desired velocity command, with preliminary simulations showing successful attainment of a stable walking gait in rough terrain. This controller was capable of capturing state relationships to determine its own linear velocity, eliminating the need for state estimation tools. For future work, we plan to integrate this policy into Unitree's B1 to conduct real-robot experiments and further assess its capabilities and robustness in challenging terrain.

%\nocite{*}
\bibliography{./format/IEEEabrv.bib, references}
\bibliographystyle{./format/IEEEtran.bst}
\end{document}